# Lattice Particle Filters

**Dirk Ormoneit**
Computer Science Dept.,
Stanford University,
Stanford, CA 94305

**Christiane Lemieux**
Mathematics and Statistics Dept.,
University of Calgary,
Alberta, Canada

**David J. Fleet**
Xerox PARC,
3333 Coyote Hill Rd,
Palo Alto, CA 94304

## Abstract

A promising approach to approximate inference in state-space models is particle filtering. However, the performance of particle filters often varies significantly due to their stochastic nature. We present a class of algorithms, called lattice particle filters, that circumvent this difficulty by placing the particles deterministically according to a Quasi-Monte Carlo integration rule. We describe a practical realization of this idea, discuss its theoretical properties, and its efficiency. Experimental results with a synthetic 2D tracking problem show that the lattice particle filter is equivalent to a conventional particle filter that has between 10 and 60% more particles, depending on their "sparsity" in the state-space. We also present results on inferring 3D human motion from moving light displays.

## 1 Introduction

The *Particle Filter* (PF) has become a popular method for approximate inference in dynamical systems, with applications such as visual tracking [1, 11, 15, 18, 20] and robot localization [8]. The PF approximates a marginal probability distribution over unknown state variables with a weighted particle set, and thereby provides a convenient approach to dealing with multimodal distributions, and nonlinear dynamics and observation equations [6, 9, 11, 14]. But despite its successes, the PF can be unstable. Statistically speaking, even though the PF produces *properly weighted* samples, the random variation of predictions based on these samples may be excessive and therefore the predictions may be unreliable. For visual tracking, this results in poor estimates of object location; in some situations it causes the algorithm to lose track of the object altogether.

This paper proposes the *Lattice Particle Filter* (LPF) as an alternative, where particles are placed deterministically according to a *lattice rule*. Lattice rules are a subclass of *Quasi-Monte Carlo* (QMC) methods, which have been used successfully for high-dimensional integration in computer graphics and finance [2, 12, 17]. An important theoretical advantage of QMC methods is that for $N$ samples the error converges at the rate $O(N^{-1} \log^s N)$, where $s$ is the state space dimension, versus $O(N^{-1/2})$ for conventional Monte Carlo (MC). From a practical viewpoint, *randomized* QMC methods can be used to construct unbiased estimators that have smaller variance than MC estimators.

After a brief introduction to the PF, we introduce QMC methods and then describe the lattice particle filter. We show quantitative results on two problems, namely, tracking 2D image patterns, and the inference of 3D human pose from a 2D binocular sequence of projected limb positions.

## 2 Previous Work

In Bayesian filtering we are interested in computing a probability distribution over the unknown state variable $\mathbf{x}_t$ at time $t$, conditioned on the observation history, $\mathbf{Y}_t \equiv (\mathbf{y}_t, ..., \mathbf{y}_1)$. This distribution, denoted $p(\mathbf{x}_t \mid \mathbf{Y}_t)$, is called the *filtering distribution*. The PF is a method for approximating $p(\mathbf{x}_t \mid \mathbf{Y}_t)$ with a set of weighted states (or particles). This approximation is updated recursively from one time step to the next as new observations become available. Gordon et al. [9] provide a clear description of the method which they call the *Bootstrap Algorithm*. In computer vision it is often called the *Condensation Algorithm* [11]. Other descriptions of the method, along with some important generalizations are given by Liu et al. [14], Doucet et al. [6], and Pitt and Shepard [19].

The goal of the particle filter is to approximate the filtering distribution $p(\mathbf{x}_t \mid \mathbf{Y}_t)$ with a set of samples. In many applications it is difficult to sample directly from $p(\mathbf{x}_t \mid \mathbf{Y}_t)$. Instead, using Bayes' rule and a Markov assumption, one can use the fact that $p(\mathbf{x}_t \mid \mathbf{Y}_t)$ is proportional to the product of a *likelihood function*, $p(\mathbf{y}_t \mid \mathbf{x}_t)$, and a *prediction distribution* that summarizes the information from previous filtering steps, $p(\mathbf{x}_t \mid \mathbf{Y}_{t-1})$. There are numerous variants of the PF that address specific shortcomings of the basic algorithm [14, 19]. In its simplest form the PF draws random proposals (states) from the prediction distribution, and evaluates the likelihood function at each proposal; normalized likelihood values serve as importance weights to account for the discrepancy between the filtering and the proposal distribution from which states were sampled.

In many applications the evaluation of the likelihood function dominates the computational cost of the algorithm. As a consequence the number of particles, and usually the dimensionality of the state space, should be kept relatively small. For example, in estimating 3D human motion, it has been useful to design low-dimensional subspaces of the state-space within which to do the filtering [20]. One can also learn better models of the body's dynamics. Of course, low dimensional representations may not exist for unconstrained motions, in which case the principal way to improve the performance of the PF is the efficient placement of particles, e.g., with importance sampling [6, 15, 19], or with MCMC sampling [3, 19].

This paper examines a Quasi-Monte Carlo (QMC) method as a general way to further reduce the number of samples. Similar approaches have only recently been suggested for filtering [7, 18]. In [18], a non-randomized QMC method is applied locally at each time step, and in [7], a form of stratified sampling with a QMC flavor is used; however, this method can only be applied in low dimensions. Here we use a *randomized* QMC method. This allows us to obtain properly weighted samples that can be used to construct unbiased estimators in the same way the PF does, and therefore a similar statistical analysis can be performed.

## 3   Bayesian Filtering

Before describing the lattice particle filter, we review the particle filter and the filtering equations. Let $\mathbf{x}_t$ denote the unknown state variable at time $t$, and let $\mathbf{y}_t$ denote the image observation at time $t$. The *transition model* characterizes the state dynamics; with a first-order Markov model it takes the form $p(\mathbf{x}_t \mid \mathbf{x}_{t-1})$. The *observation model* specifies the probabilistic relation between the state and the observations, providing a *likelihood function*, $p(\mathbf{y}_t \mid \mathbf{x}_t)$. Then, the filtering dis-

tribution $p(\mathbf{x}_t \mid \mathbf{Y}_t)$ is given by

$$
\begin{aligned}
p(\mathbf{x}_t | \mathbf{Y}_t) &= \int ... \int p(\mathbf{x}_t, ..., \mathbf{x}_0 | \mathbf{Y}_t) d\mathbf{x}_1 ... d\mathbf{x}_{t-1} \\
&\propto \int ... \int \prod_{k=1}^{t} p(\mathbf{y}_k | \mathbf{x}_k) p(\mathbf{x}_k | \mathbf{x}_{k-1}) d\mathbf{x}_k \;. \quad (1)
\end{aligned}
$$

The direct evaluation of (1) is difficult in practice because it involves an integral whose dimension grows with $t$. Hence it is advisable to solve (1) recursively using the *prediction* and *filtering* equations:

$$
\begin{aligned}
p(\mathbf{x}_t | \mathbf{Y}_{t-1}) &= \int p(\mathbf{x}_t | \mathbf{x}_{t-1}) p(\mathbf{x}_{t-1} | \mathbf{Y}_{t-1}) d\mathbf{x}_{t-1} \quad (2) \\
p(\mathbf{x}_t | \mathbf{Y}_t) &\propto p(\mathbf{y}_t | \mathbf{x}_t) p(\mathbf{x}_t | \mathbf{Y}_{t-1}) \;. \quad (3)
\end{aligned}
$$

If the transition and observation models are linear with Gaussian noise, then (2) and (3) are also Gaussian and can be updated using the Kalman filter [10]. Otherwise, the representation and computation of (2) and (3) can be difficult. In this case, one approach is to approximate $p(\mathbf{x}_t | \mathbf{Y}_t)$ with a weighted set of particles,

$$
\mathcal{S}_t = \{ (\mathbf{x}_{t,i}, w_{t,i}) \}_{i=1}^{N} \;. \quad (4)
$$

This idea can be made precise by requiring that $\mathcal{S}_t$ be *properly weighted* [14] in the sense that

$$
\sum_{i=1}^{N} w_{t,i} f(\mathbf{x}_{t,i}) \stackrel{N \to \infty}{\to} E[f(\mathbf{x}_t | \mathbf{Y}_t)] \quad (5)
$$

for arbitrary integrable functions $f$.

The PF generates properly weighted samples [14] and it encourages the exploration of paths $(\mathbf{x}_{1,i}, ..., \mathbf{x}_{t,i})$ that appear likely given $\mathbf{Y}_t$, discarding particles with small weights from (4). The PF is described by:

$$
\begin{aligned}
a(i) &\sim \text{multinomial}(w_{t-1,1}, ..., w_{t-1,N}) \;, \quad (6) \\
\mathbf{x}_{t,i} &:= g(\mathbf{u}_{t,i}, \mathbf{x}_{t-1,a(i)}) \;, \quad (7) \\
w_{t,i} &:= \frac{p(\mathbf{y}_t \mid \mathbf{x}_{t,i})}{\sum_{j=1}^{N} p(\mathbf{y}_t \mid \mathbf{x}_{t,j})} \;. \quad (8)
\end{aligned}
$$

The index sequence generated in step (6) specifies the "surviving" particles, $\mathbf{x}_{t-1,a(i)}, i = 1, ..., N$. These particles are propagated forward in step (7), and step (8) specifies the weights associated with each new particle. The points $\mathbf{u}_{t,i}$ are independent, and $g(\mathbf{u}_{t,i}, \mathbf{x}_{t-1,i})$ is a *transformation function* that maps the uniform vector $\mathbf{u}_{t,i}$ and the previous state $\mathbf{x}_{t-1,i}$ onto an sample from $p(\mathbf{x}_t | \mathbf{x}_{t-1,i})$. This transformation is a component of any computer simulation since random variate generation for many continuous distributions are based on transformations of uniform random numbers [5]. In the case of a univariate standard normal transition model, $g(\mathbf{u}_{t,i}, 0)$ would be the inverse standard normal distribution function.

The LPF provides an improved way of executing the filtering steps (6) − (7). The idea is to draw points $\mathbf{u}_{t,i}$ from a lattice rule instead of using random sampling, as we explain in the next section.

## 4   Lattice Particle Filter

The PF described in Sec. 3 produces properly weighted samples that approximate the filtering distribution (1). However, the PF's performance may vary considerably in practice due to the random nature of the sampling. This is sometimes particularly harmful in the context of filtering because, even though the approximation error due to sampling at each step may be small, errors can accumulate over time in an exponential manner.

To see this, consider a simple example where the transition density $p(\mathbf{x}_t|\mathbf{x}_{t-1})$ is uniform on the state-space $[0,1)$ and independent of $\mathbf{x}_{t-1}$. Let the observations be binary with $P(\mathbf{y}_t = 1|\mathbf{x}_t)$, being 1 if $\mathbf{x}_t < 0.2$ and 0 otherwise. That is, the system states $\mathbf{x}_t$ are placed independently on the unit interval, but only those states in $[0, 0.2)$ trigger the response $\mathbf{y}_t = 1$. Suppose furthermore that the *true* state trajectory of $\mathbf{x}_t$ happened to evolve entirely within $[0, 0.2)$ so that $\mathbf{y}_t = 1$ for all $t$. And assume we use a PF with $N = 10$ particles to recover the trajectory. At each time step the number of particles within $[0, 0.2)$ is binomially distributed with parameters 10 and 0.2. Hence the probability of having no particles within $[0, 0.2)$ in any one of $k$ time steps is $(1 - (1 - 0.8^{10})^k)$, which converges to one exponentially with $k$. In other words, after a sufficiently long time there almost certainly occurs one time where the PF lost track of the object completely.

In this example, an alternative strategy would have been to choose only one particle at random at each time, and to place the remaining particles equidistantly around that particle. Then there would always be two particles in $[0, 0.2)$. This deterministic placement of particles according to a *low-discrepancy rule* is a special case of a QMC method. We next provide some general background on QMC methods and lattice rules, and then describe the LPF.

### 4.1   Quasi-Monte Carlo Methods

Consider a generic integration problem with respect to the uniform measure over the domain $[0,1)^s$. There are several ways to approach this task. MC integration averages the values of the integrand at $N$ random, uniform, independent points in $[0,1)^s$. Its error has a convergence rate of $O(N^{-\frac{1}{2}})$ that is independent of $s$. In many practical problems, this convergence rate is too slow. A deterministic alternative might be to use a Cartesian product of one-dimensional point sets, as shown in Figure 1(left) for $s = 2$, but with this deterministic approach the number of points $N$ needed to preserve a constant level of integration accuracy grows exponentially fast with the dimension $s$. This suboptimal behavior is related to the fact that such constructions have low-dimensional projections containing less than $N$ points; for example, note that in Figure 1(left), the 256 points project to only 16 points along each axis to integrate the $x_1$ and $x_2$ components of an integrand of the form $f(x_1, x_2)$.

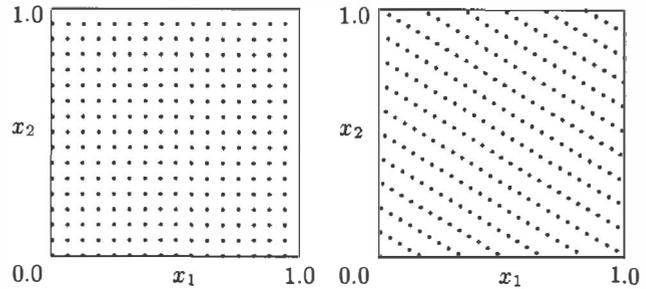

Figure 1: *Left:* Rectangle rule. *Right:* Korobov lattice rule with $N = 256$ and $a = 25$; the projection of the points on each axis results in $N$ distinct values.

By contrast, QMC methods aim to create point sets whose different projections onto coordinate subspaces always contain $N$ distinct points. An overview of random number generation and QMC methods is given in the book by Niederreiter [16]. An example of a 2D QMC point set, called a Korobov lattice rule, is shown in Figure 1(right).

A second goal of QMC methods is to create point sets that are as close as possible to a uniform distribution. More precisely, QMC methods are based on so-called *low-discrepancy point sets* which give convergence rates of $O(N^{-1} \log^s N)$ for integration error, in contrast to the MC method that is $O(N^{-\frac{1}{2}})$. In practice, the performance of QMC integration is intimately related to the *effective dimension* of the integration problem, i.e., the ability to approximate the integrand by a sum of low-dimensional functions [2, 17]. Integrands having a small effective dimension can be integrated accurately by QMC point sets that have good projections over subspaces of low dimension. Superior performance of QMC over MC has been demonstrated in numerous applications [2, 12, 17].

As a special case of QMC methods we consider *lattice rules* here [13, 21]. By comparison to the Sobol and Niederreiter sequences [16] used in [18], lattice rules are easy to implement (only one parameter, the *generator*, is required for a given sample size), and can be built so that their projections onto low-dimensional subspaces are not only well distributed but also of equivalent quality for different subspaces. They can also be randomized to produce unbiased estimators.

| $\log_2 N$ | 4 | 5 | 6 | 7 | 8 |
|---|---|---|---|---|---|
| $a_l$ | 3 | 5 | 11 | 13 | 25 |
| $a_h$ | 3 | 5 | 5 | 11 | 75 |
| $\log_2 N$ | 9 | 10 | 11 | 12 | 13 |
| $a_l$ | 55 | 43 | 259 | 307 | 699 |
| $a_h$ | 51 | 139 | 519 | 1081 | 1289 |
| $\log_2 N$ | 14 | 15 | 16 | 17 | 18 |
| $a_l$ | 2087 | 7243 | 11035 | 27891 | 18373 |
| $a_h$ | 2961 | 2149 | 21553 | 27383 | 3597 |
| $\log_2 N$ | 19 | 20 | 21 | | |
| $a_l$ | 21643 | 201579 | 431119 | | |
| $a_h$ | 120079 | 172565 | 232501 | | |

Table 1: Generators for state-space dimensions up to 8 should use generators, $a_l$. Problems with dimensions up to 32 use generator, $a_h$.

To build the LPF, we use a special construction of lattice rule called a *shifted Korobov lattice rule* based on $N$ points $\mathbf{U}_1, \ldots, \mathbf{U}_N$ given by

$$\mathbf{U}_i = \left( \frac{i-1}{N}(1, a, \ldots, a^{s-1}) + \boldsymbol{\Delta} \right) \bmod 1 , \quad (9)$$

for $i = 1, \ldots, N$, where the modulo 1 is applied component-wise. The integer $a \in [1, ..., N-1]$ is called the *generator* of the rule; its choice is crucial for the performance. Suitable values for $a$, depending on the sample size $N$ and the state dimension, are given in Table 1 (see [13] for more details). The vector $\boldsymbol{\Delta}$ is a shift that is uniformly distributed over $[0, 1)^s$, which implies that the points $\mathbf{U}_i$ are also uniform, but *dependent*. This shift is important as a means of obtaining error estimates via multiple simulations [4], and it is necessary to guarantee that the resulting approximations are *unbiased*.

### 4.2  LPF Algorithm

The LPF algorithm propagates $N$ particles following the rules (6), (7), and (8). However, the uniform numbers $\mathbf{u}_{t,i}$ in the forward propagation (7) become components of the (shifted) lattice rule (9).

More precisely, we start with a shifted lattice rule $P_N = \{\mathbf{U}_i, i = 1, \ldots, N\}$ in $sT$ dimensions, where $s$ is the dimension of the state at a single time, and $T$ is the number of time steps. We decompose $P_N$ into components of the form $\mathbf{u}_{t,1}, \mathbf{u}_{t,2}, \ldots, \mathbf{u}_{t,N}$ for each time $t$. Note that, due to (9), these components can be computed recursively for $t = 1, \ldots, T$, without storing $P_N$ explicitly. Next, we use $\mathbf{u}_{t,1}, \mathbf{u}_{t,2}, \ldots, \mathbf{u}_{t,N}$ for forward propagation in step (7). An important issue is the assignment of a particular uniform number $\mathbf{u}_{t,j}$ to the uniform variables $\mathbf{u}_{t,i}$ in (7). There is a danger of assigning components to particles that might depend on the outcome of the resampling, hence introducing a serious bias. Our approach is to generate at each time $t$ a uniform and random permutation $\Gamma_t = (\gamma_{t,1}, \ldots, \gamma_{t,N})$ of the integers $[1 \ldots N]$, and then assign the point $\mathbf{u}_{t,\gamma_{t,i}}$ to particle $i$. This implies that the trajectory $\mathbf{x}_{1,i}, \ldots, \mathbf{x}_{t,i}$ of the $i^{th}$ particle from time 1 to $t$ is generated by the point

$$\tilde{\mathbf{U}}_{t,i} = (\mathbf{u}_{1,\gamma_{1,i}}, \mathbf{u}_{2,\gamma_{2,i}}, \ldots, \mathbf{u}_{t,\gamma_{t,i}}). \quad (10)$$

Note that the point set $\{\tilde{\mathbf{U}}_{t,1}, \ldots, \tilde{\mathbf{U}}_{t,N}\}$ does not form a lattice rule, which would be necessary to obtain the $O(N^{-1} \log^s N)$ convergence rate *globally*. However, the point set used at each time $t$ forms a lattice rule. In addition, it can be shown [13] that when $\gcd(a, N) = 1$, the point sets used at each time step differ in the implementation only because of the random shift. Hence the quality of the sampling remains the same throughout time, which is generally not the case for other types of QMC point sets such as those used in [18].

In summary, the LPF implementation only differs from the PF in the use of the points $\mathbf{u}_{t,i}$ used to generate the state $\mathbf{x}_{t,i}$. Instead of calling a pseudorandom number generator $s$ times to define $\mathbf{u}_{t,i}$, the LPF instead generates *one* random shift $\Delta$, one random permutation $\Gamma_t = (\gamma_{t,1}, \ldots, \gamma_{t,N})$ of the integers from 1 to $N$, and then use the point

$$\left( \frac{\gamma_{t,i} - 1}{N}(1, a, \ldots, a^{s-1}) + \Delta \right) \bmod 1$$

to generate the state $\mathbf{x}_{t,i}$ of the $i^{th}$ particle.

## 5  Theoretical Aspects

In this section we discuss some theoretical results and we outline differences and similarities between the LPF and the ordinary PF. In terms of search methods, one can view the ordinary PF as an informed, random search where different nodes of a search tree are expanded or truncated at different time steps. Conversely, the LPF corresponds to an (almost) deterministic search algorithm. The observation model is used as a heuristic function to determine which "node" (i.e. branch of the filtering tree) is expanded next, and the transition model specifies how to carry out this expansion. In the LPF the expansion is designed to search the state space as *evenly* as possible.

From a mathematical perspective, we can show that the samples generated by the LPF are properly weighted in the sense of (5). This is suggested by the fact that the updating of the weights $w_{t,i}$ and the sampling with replacement in the LPF are done exactly as in the PF. The only difference is that the particles $\mathbf{x}_{t,1}, \ldots, \mathbf{x}_{t,N}$ are not independent. We formalize this intuition in the following theorem, which covers the LPF as a special case:

**Theorem 1** *If each particle $\mathbf{x}_{t,i}$ is generated by a point $\mathbf{u}_{t,i}$ that has a uniform distribution over $[0,1)^s$ and such that $\mathbf{u}_{t,i}$ and $\mathbf{u}_{r,j}$ are independent for $1 \leq r < t \leq T$, $1 \leq i,j \leq N$, and if the resampling step at time $t$ is done by sampling with replacement using weights $w_{t,i}$ proportional to $p(\mathbf{y}_t|\mathbf{x}_{t,i})$, then $\mathcal{S}_t = \{\mathbf{x}_{t,i}, w_{t,i}\}_{i=1}^N$ is a properly weighted sample, for $t = 1, \ldots, T$.*

The proof is omitted due to space limitations; it proceeds by showing that each $\mathbf{x}_{t,i}$ follows the distribution $p(\mathbf{x}_t|\mathbf{Y}_{t-1})$ as in (2), and then, that $E[f(\mathbf{x}_{t,i})\,p(\mathbf{y}_t|\mathbf{x}_{t,i})] \propto E[f(\mathbf{x}_t|\mathbf{Y}_t)]$ for any function $f$; from there, applying the ergodic theorem yields the result. Hence, at least asymptotically, the LPF produces a correct approximation of the filtering distribution (3), and furthermore, the quantity $\sum_{i=1}^N w_{t,i}f(\mathbf{x}_{t,i})$ is an unbiased estimator of $E[f(\mathbf{x}_t|\mathbf{Y}_t)]$. Note that the random shift in (9) is essential to this result and is a principal difference between the LPF and previous work on QMC-based filtering where samples are generated deterministically, e.g. [18].

# 6 Experiments

To test the LPF we compare it to a conventional PF that uses residual sampling [14]. Both filters had the same observations, the same temporal dynamics, and the same likelihood function. Our goal is to compare the different filters in how well they approximate the filtering distribution (3). The quality of the approximation is measured by computing the error covariation in the estimation of the mean of the filtering distribution across many runs of the filters. By applying the filters with varying numbers of particles we can also analyze the relationship between estimation errors and the computation time required by both algorithms. Computational requirements in both algorithms scale similarly with the number of particles.

## 6.1 Disk Tracking

The first experiment involves a circular disk undergoing a random walk in an image sequence. The disk position at time $t$, $\mathbf{x}_t \in \mathbb{R}^2$, is given by

$$\mathbf{x}_t = \mathbf{x}_{t-1} + \eta_t , \qquad (11)$$

where $\eta_t$ is 2D i.i.d. zero-mean Gaussian noise; i.e., $\eta_t \sim N(\bar{0}, \sigma_x^2 \mathcal{I}_2)$, where $\mathcal{I}_2$ is the $2 \times 2$ identity matrix. The standard deviation was $\sigma_x = 3$ pixels.

The effective scene model $m(\mathbf{r},t)$ is 1 if pixel location $\mathbf{r}$ is within a disk of radius 16 pixels, centered at $\mathbf{x}_t$:

$$m(\mathbf{r},t) \equiv \begin{cases} 1 & \text{for } \|\mathbf{r} - \mathbf{x}_t\| \leq 16 \\ 0 & \text{otherwise .} \end{cases} \qquad (12)$$

The image observation is obtained by adding i.i.d. zero-mean Gaussian noise $\nu_{t,\mathbf{r}} \sim N(0, \sigma_\nu^2)$, with $\sigma_\nu =$

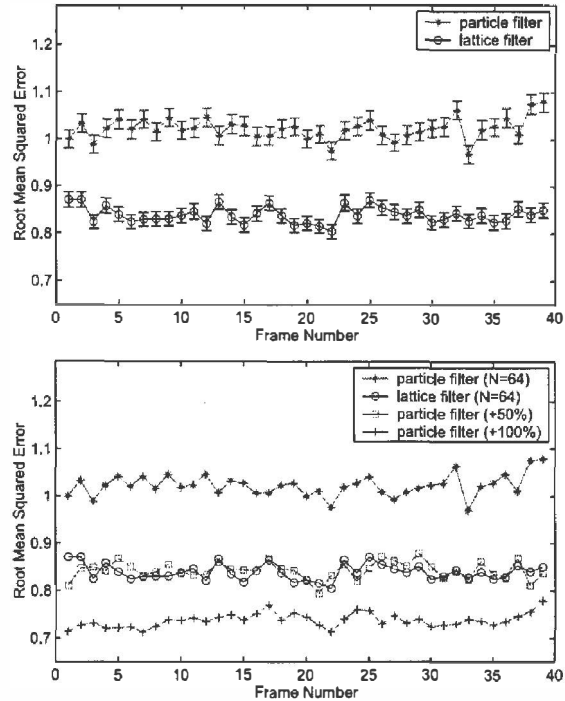

Figure 2: Root mean-squared errors (RMSE) in the estimated disk location for a PF with residual sampling and for the LPF. (TOP) RMSE with confidence levels based on 64 samples. (BOTTOM) RMSE for PF's with different sample sizes; with 50% more samples, the PF produces errors similar in magnitude to the LPF.

0.25, to $m(\mathbf{r},t)$:

$$\mathbf{I}(\mathbf{r},t) \equiv m(\mathbf{r},t) + \nu_{t,\mathbf{r}} . \qquad (13)$$

The model transition density is

$$p(\mathbf{x}_t|\mathbf{x}_{t-1}) \equiv N(\mathbf{x}_t|\mathbf{x}_{t-1}, \sigma_d^2 \mathcal{I}_2) . \qquad (14)$$

We misspecified the model (14) by setting $\sigma_d = 5$, which is about 60% larger than the standard deviation in the "true" transition model (11). This is because such model parameters are typically unknown in practice, and hence a relatively large value must be chosen to search a sufficiently large neighborhood of $\mathbf{x}_t$.

We created 1000 image sequences, each with 40 frames, by simulating (11), (12), and (13). We applied the PF and the LPF to each sequence, initialized by setting $\mathbf{x}_0$ to the true disk location at time 0. The disk location was estimated by taking the weighted mean of the particles, according to (5). The error at each time step is the Euclidean distance between the estimated and the true disk location. As a summary statistic, we compute the root mean-squared estimation error (RMSE) of the 1000 trials at each time step.

The RMSE results of an experiment with 64 particles are shown in Figure 2 as a function of time. The cir-cles represents the RMSE of the LPF and the other curves represent the performance of the ordinary PF for different numbers of particles. Figure 2 (top) compares the LPF to the PF with identical numbers of particles, showing RMSE and standard error bars that provide a measure of our confidence in this expected deviation from the true disk location. The LPF produces location errors that are approximately 20% less than the PF with the same number of samples. This supports our hypothesis that the deterministic placement of particles improves the average performance of the estimate. In addition, although difficult to discern from this plot, the standard error bars from the LPF are also smaller.

| # particles | 16 | 32 | 64 | 128 | 256 | 512 |
|---|---|---|---|---|---|---|
| Var difference | 20% | 21% | 19% | 11% | 11% | 10% |
| efficiency gain | 50% | 60% | 50% | 15% | 20% | 20% |

Table 2: Experimental results in disk tracking task

Figure 2 (bottom) shows the same RMSE results for the LPF but compared against RMSE results for PFs with different numbers of particles. This allows one to assess the computational gains due to the LPF. One can see that we need to increase the number of particles, and hence the computation time, for the PF by approximately 50% in order to produce errors as small as those obtained with the LPF. Even with 50% more particles the performance of each individual trial may still be worse for the PF due to the larger error bars.

These results hold over a wide range of particle sizes. Figure 3 shows other experiments with from 16 particles up to 512 particles. The experimental setup is identical to that in Figure 2. Note that the average prediction error of both methods, as well as the variability of the predictions, decreases with an increasing number of particles. A summary of the results from the complete set of experiments is given in Table 2.

Note that the LPF consistently outperforms the PF by a margin of at least 15%. The performance improvement seems to be less pronounced in the cases where we have relatively many particles (128, 256, and 512). This is to be expected because an optimized particle placement seems particularly relevant in the case where there are relatively few particles. Regarding the percentage of additional particles needed by the PF to match the performance of the LPF, which is reported in the third row of the table, the maximum difference of 60% occurs at a sample size of 32 particles for the LPF. Again, this difference becomes less pronounced as the overall number of particles increases.

## 6.2   Human Motion Tracking

In our second set of experiments, we apply both the LPF and the PF using residual sampling to a human motion tracking problem. The task is to recover the lower portion of a human body from 2D projections. Rather than using camera images as observations, we used 2D projections of body markers (on the joints) obtained from a commercial motion capture system. This filtering problem is challenging because the lower body (legs and hips) is described in a 10D space. Tracking occurs directly on this space without first applying algorithms for dimensionality reduction like principal component analysis [20].

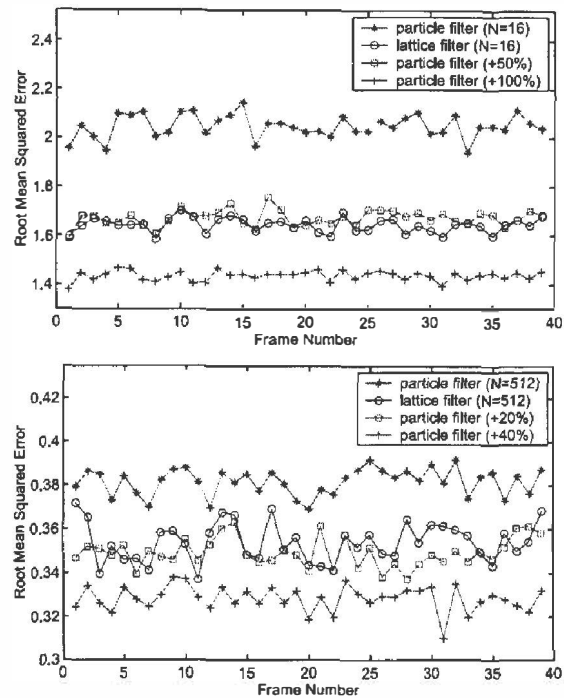

Figure 3: Root mean-squared errors (RMSE) in the estimated disk location for PF's with residual sampling and for the LPF. The number of samples for the different filters are show in the legends in each plot.

The image observations are labeled 2D positions corresponding to markers on a human subject, the 3D locations of which were found with a commercial motion capture system. The 6 markers used for tracking the lower body are shown in Fig. 4. The observation model involves the perspective projection of the 3D points onto the image plane plus additive Gaussian noise. Given a camera center at $(X_c, Y_c, Z_c)$, with the optical axis parallel to the Y-axis, a marker point $(X_m, Y_m, Z_m)$ produces the observation

$$\mathbf{d} = \left( \frac{X_m - X_c}{Y_m - Y_c}, \frac{Z_m - Z_c}{Y_m - Y_c} \right) + \eta \qquad (15)$$

where $\eta$ is isotropic 2D mean-zero Gaussian noise with variance $\sigma^2$. In the experiments below we used $\sigma = 0.002$, which is approximately 2% the length of an upright spine projected into the image plane.

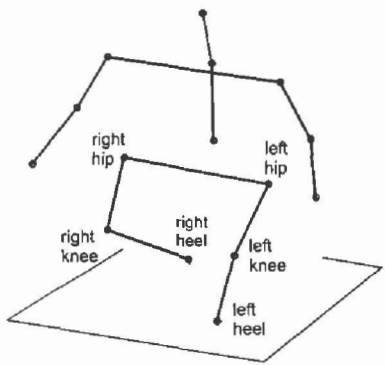

Figure 4: Locations of 6 markers on the lower body.

The body is modeled as an articulated linkage. The state vector for the lower body model has 10 angular degrees of freedom (DOF): 2 for the pelvis, 3 for each hip, and 1 for each knee. Given a state, $\mathbf{x}$, and the known spine location, one can traverse the kinematic tree, from the spine to the pelvis, and then to each leg, to generate the 3D marker locations. These 3D locations are then projected into the image to generate predicted observations from the model. Let $\hat{\mathbf{d}}_j(\mathbf{x})$ be the 2D location for the $j^{th}$ marker that is predicted by state $\mathbf{x}$. Then the likelihood function for state s and observations $D = \{\mathbf{d}_j\}_{j=1}^6$ is

$$p(D \mid \mathbf{x}) = \left(\frac{1}{2\pi\sigma^2}\right)^6 \exp\left[-\frac{\sum_{j=1}^6 ||\mathbf{d}_j - \hat{\mathbf{d}}_j(\mathbf{x})||^2}{2\sigma^2}\right] , (16)$$

where $\sigma^2$ is the variance of the image noise.

The experiments simulate a binocular observer that views the person from approximately 2.5m, with 6cm between the two eyes. Assuming independent noise in the two image views, the joint likelihood is the product of the individual likelihoods. For a temporal prior over these state variables, we make the simplistic assumption that the state changes slowly over time. In particular, the transition density for states $\mathbf{x}_t$, conditioned on $\mathbf{x}_{t-1}$, is an isotropic, mean-zero Gaussian, with a variance of $\sigma_a^2$. Here, we fix $\sigma_a = 0.1$ radians.

Note that we do not impose limits on how far joints can rotate. The tracker may draw high-likelihood proposals that are anatomically impossible, but are consistent with the 2D observations. Also, limb lengths are determined from the dimensions of the human subject before tracking begins and are then fixed during tracking. This is a source of error as real joints are more complicated than our model, causing some parts such as the pelvis to vary in width over time.

Comparing the performance of the LPF and PF is more difficult in this case. In particular, the mean of the filtering distribution is not always equal to the true state. As the filters are attempting to approximate the filtering distribution, an appropriate mea-sure of performance is the difference between the true mean of the filtering distribution and the mean states computed using the approximations provided by the LPF and the PF.

To obtain a ground truth measure of the true mean of the filtering distribution, we ran a PF on the input sequences with 16 times the number of samples than we used in the experiments. We then took the true mean to be the sample mean from this large run. We then ran the LPF and the PF 200 times on the same input, with different random seeds on every run. The means on each run were computed according to (5).

Figure 5 summarizes the results for individual state variables of the left leg, namely, two angles from the left hip, and one for the left knee. Notice from Fig. 5 (top) that the average of the mean estimates obtained over the 200 trials is very close to the true mean in both cases. This is not surprising since both the LPF and the PF produced unbiased estimates. The error bars in Fig. 5 (top) show the standard deviation of the ensemble of means for the 200 runs, which is where we expect the LPF to show smaller variability of the estimates about the true mean. Like the disk tracker results, the standard deviation for the LPF is usually 5% - 20% below that for the particle filter, with similar gains in computational efficiency.

Figure 5 (bottom) shows the expected (absolute) difference between the true mean and the mean estimates obtained with the LPF and the PF. The error bars are larger than in Fig. 2 because we have only 200 instead of 1000 trials. The lower errors with the lattice method are clear nonetheless.

# 7   Conclusions

We presented an algorithm called the lattice particle filter to improve the reliability of particle filters in visual tracking. Specifically, the method places the particles deterministically according to a randomly shifted lattice rule. This reduces the variance of the particle-based approximation. Experiments demonstrate that the practical improvement from using the LPF amounts to savings in the number of particles of between 20% and 60%. The size of this effect depends on the "sparsity" of the particles in the state-space, being more pronounced when there are relatively few particles as it is typical for real applications. In our experiments involving a 10D human motion tracker, the lattice particle filter also led to similar reductions in variance.

For future work, we plan alternative versions of the LPF that are based on a global lattice rule in $nT$ dimensions. We believe that the global lattice property will be central to obtain exact convergence rates and to optimize the practical performance of the LPF. However, it involves the creation of particles satisfying a global lattice constraint. We will also explore alternative resampling methods that could, e.g., depend on the variability of the likelihoods obtained at each time step, such as those in [6, 14].

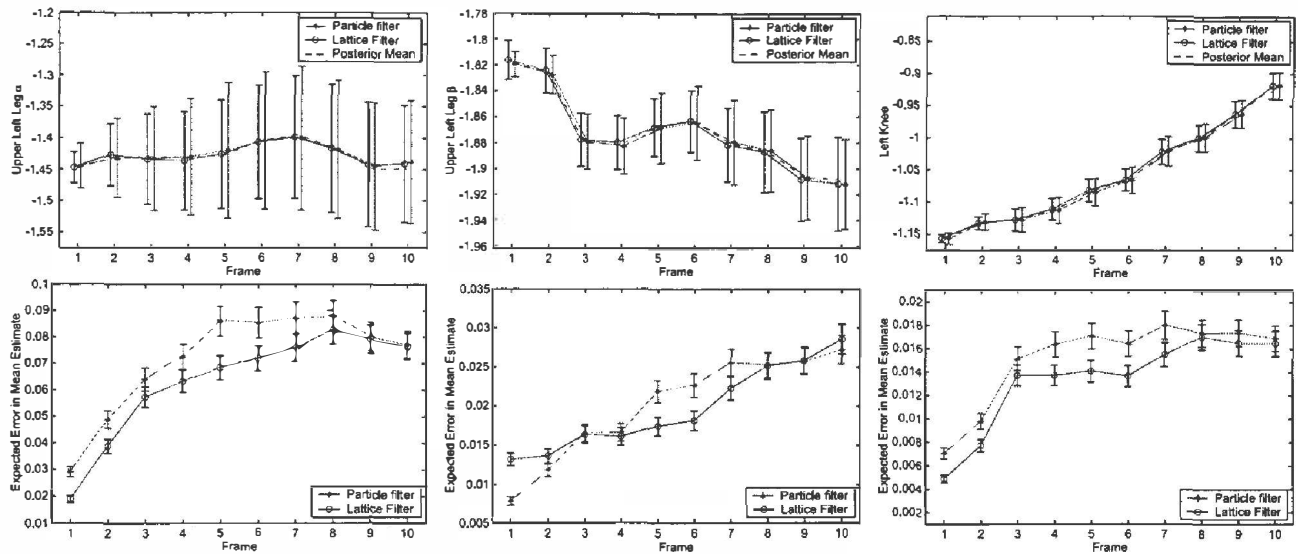

Figure 5: Summary of 200 trials of PF and LPF tracking of lower body. (Top) Individual state variables for the left leg, for which the LPF means are generally is closer to the posterior mean, and have significantly less variance. (Bottom) Expected difference between the posterior mean and the means computed by the PF and the LPF.